\documentclass[11pt,a4paper]{article}
\usepackage[hyperref]{acl2020}
\usepackage{times}
\usepackage{latexsym}

\usepackage{microtype}

\usepackage{graphicx}
\usepackage{amsmath}
\usepackage{amsfonts}
\usepackage{multirow}
\usepackage{subcaption}
\aclfinalcopy 
\def\aclpaperid{3193} 

\title{Gated Convolutional Bidirectional Attention-based Model for \\ Off-topic Spoken Response Detection}

\author{
Yefei~Zha$^1$~~~~Ruobing~Li$^1$~~~~Hui~Lin$^{1,2}$\\\\
$^1$LAIX Inc.\\
$^2$Shanghai Key Laboratory of Artificial Intelligence in Learning and Cognitive Science \\\\
\texttt{\{leo.zha, ruobing.li, hui.lin\}@liulishuo.com} 
}

\date{}

\begin{document}
\maketitle
\begin{abstract}
Off-topic spoken response detection, the task aiming at predicting whether a response is off-topic for the corresponding prompt, is important for an automated speaking assessment system.
In many real-world educational applications, off-topic spoken response detectors are required to achieve high recall for off-topic responses not only on seen prompts but also on prompts that are unseen during training.
In this paper, we propose a novel approach for off-topic spoken response detection with high off-topic recall on both seen and unseen prompts.
We introduce a new model, Gated Convolutional Bidirectional Attention-based Model (GCBiA),
which applies bi-attention mechanism and convolutions to extract topic words of prompts and key-phrases of responses, and introduces gated unit and residual connections between major layers to better represent the relevance of responses and prompts.
Moreover, a new negative sampling method is proposed to augment training data.
Experiment results demonstrate that our novel approach can achieve significant improvements in detecting off-topic responses with extremely high on-topic recall, for both seen and unseen prompts.
\end{abstract}

\section{Introduction}
Off-topic spoken response detection is a crucial task in an automated assessment system. 
The task is to predict whether the response is off-topic for the corresponding question prompt.
Table~\ref{tab:example-off-topic} shows an example of on-topic and off-topic responses for a prompt.

Off-topic examples in human-rated data is often too sparse to train an automated scoring system to reject off-topic responses.
Consequently, automated scoring systems tend to be more vulnerable than human raters to scoring inaccurately due to off-topic responses (~\citealp{lochbaum2013detection};~\citealp{higgins2014managing}).
To ensure the validity of speaking assessment scores, it is necessary to have a mechanism to flag off-topic responses before scores are reported~\citep{wang2019automatic}.
In our educational application, we use the automated speaking assessment system to help L2 learners prepare for the IELTS speaking test. 
We do see a higher rate of off-topic responses in freemium features as some users just play with the system. 
In such a scenario, accurate off-topic detection is extremely important for building trust and converting trial users to paid customers.

\begin{table}[h!]
\centering
\begin{tabular}{p{7cm}}
\hline
\textbf{Prompt:} \textit{What kind of \textbf{flowers} do you like?} \\
\textbf{On-topic:} I like \textbf{\textit{iris}} and it has different meaning of it a wide is the white and um and the size of a as a ride is means the ride means love but I can not speak. \\
\textbf{Off-topic:} Sometimes I would like to invite my friends to my home and we can play the Chinese chess dishes this is my favorite games at what I was child. \\
\hline
\end{tabular}
\caption{An example of on-topic and off-topic responses for a prompt.}
\label{tab:example-off-topic}
\end{table}

Initially, many researchers used vector space model (VSM) (~\citealp{louis2010off};~\citealp{yoon2014similarity};~\citealp{evanini2014automatic}) to assess the semantic similarity between responses and prompts.
In recent years, with the blooming of deep neural networks (DNN) in natural language processing (NLP), many DNN-based approaches were applied to detect off-topic responses.
~\citet{malinin2016off} used the topic adapted Recurrent Neural Network language model (RNN-LM) to rank the topic-conditional probabilities of a response sentence. 
A limitation of this approach is that the model can not detect off-topic responses for new question prompt which was not seen in training data ({\em unseen prompt}).
Later, off-topic response detection was considered as a binary classification task using end-to-end DNN models.
~\citet{malinin2017attention} proposed the first end-to-end DNN method, attention-based RNN (Att-RNN) model, on off-topic response detection task. 
They used a Bi-LSTM embedding of the prompt combined with an attention mechanism to attend over the response to model the relevance.
CNNs may perform better than RNNs in some NLP tasks which require key-phrase recognition as in some sentiment detection and question-answer matching issues~\citep{yin2017comparative}.
~\citet{lee2017off} proposed a siamese CNN to learn semantic differences between on-topic response-questions and off-topic response-questions.
~\citet{wang2019automatic} proposed an approach based on similarity grids and deep CNN.

However, the cold-start problem of off-topic response detection has not been handled well by the aforementioned approaches.
It is not until enough training data of unseen prompts are accumulated that good performance could be achieved.
Besides, these methods draw little attention to the vital on-topic false-alarm problem for a production system. I.e., extremely high recall of on-topic responses is also required to make real-user-facing systems applicable.


In this paper, to address the issues mentioned above, a novel approach named Gated Convolutional Bidirectional Attention-based Model (GCBiA) and a negative sampling method to augment training data are proposed.
The key motivation behind our model GCBiA is as follows: convolution structure captures the key information, like salient n-gram features~\citep{young2018recent} of the prompt and the response, while the bi-attention mechanism provides complementary interaction information between prompts and responses.
Following R-Net~\citep{wang2017gated} in machine comprehension, we add the gated unit as a relevance layer to filter out the important part of a response regarding the prompt.
These modules contribute to obtaining better semantic matching representation between prompts and responses, which is beneficial for both seen and unseen prompts.
Additionally, we add residual connections~\citep{he2016deep} in our model to keep the original information of each major layer.
To alleviate the cold-start problem on unseen prompts, a new negative sampling data augmentation method is considered.

We compare our approach with Att-RNN model and G-Att-RNN (our strong baseline model based on Att-RNN).
Experiment results show that GCBiA outperforms these methods both on seen and unseen prompts benchmark conditioned on extremely high on-topic response recall (0.999). 
Moreover, the model trained with negative sampling augmented data achieves 88.2 average off-topic recall on seen prompts and 69.1 average off-topic recall on unseen prompts, respectively.

In summary, the contribution of this paper is as follows:
\begin{itemize}
\item We propose an effective model framework of five major layers on off-topic response detection task. The bi-attention mechanism and convolutions are applied to the focus on both topic words in prompts and key-phrase in responses. The gated unit as a relevance layer can enhance the relevance of prompts and responses. Besides, residual connections for each layer were widely used to learn additional feature mapping.
Good semantic matching representation is obtained by these modules on both seen and unseen prompts.
The GCBiA model achieves significant improvements by +24.0 and +7.0 off-topic recall on average unseen and seen prompts respectively, comparing to the baseline method.
\item To explore the essence of our proposed model, we conduct visualization analysis from two perspectives: bi-attention visualization and semantic matching representation visualization to reveal important information on how our model works.
\item To improve our result on unseen prompts further, we propose a novel negative sampling data augmentation method to enrich training data by shuffling words from the negative sample in off-topic response detection task. It allows the GCBiA model to achieve higher averaging off-topic recall on unseen prompts.
\end{itemize}

\section{Approach}
\subsection{Task formulation}
The off-topic response detection task is defined as follows in this paper. Given a question prompt with $n$ words $X^P=\{x^P_t\}^n_{t=1}$ and the response sentence with $m$ words $X^R=\{x^R_t\}^m_{t=1}$, output one class $o=1$ as on-topic or $o=0$ as off-topic.

\subsection{Model Overview}
We propose a model framework of five major layers on off-topic response detection task. 
The proposed model GCBiA (shown in Figure~\ref{fig:model}) consists of the following five major layers:
\begin{figure*}[h!]
\centering
\includegraphics[width=0.88\textwidth]{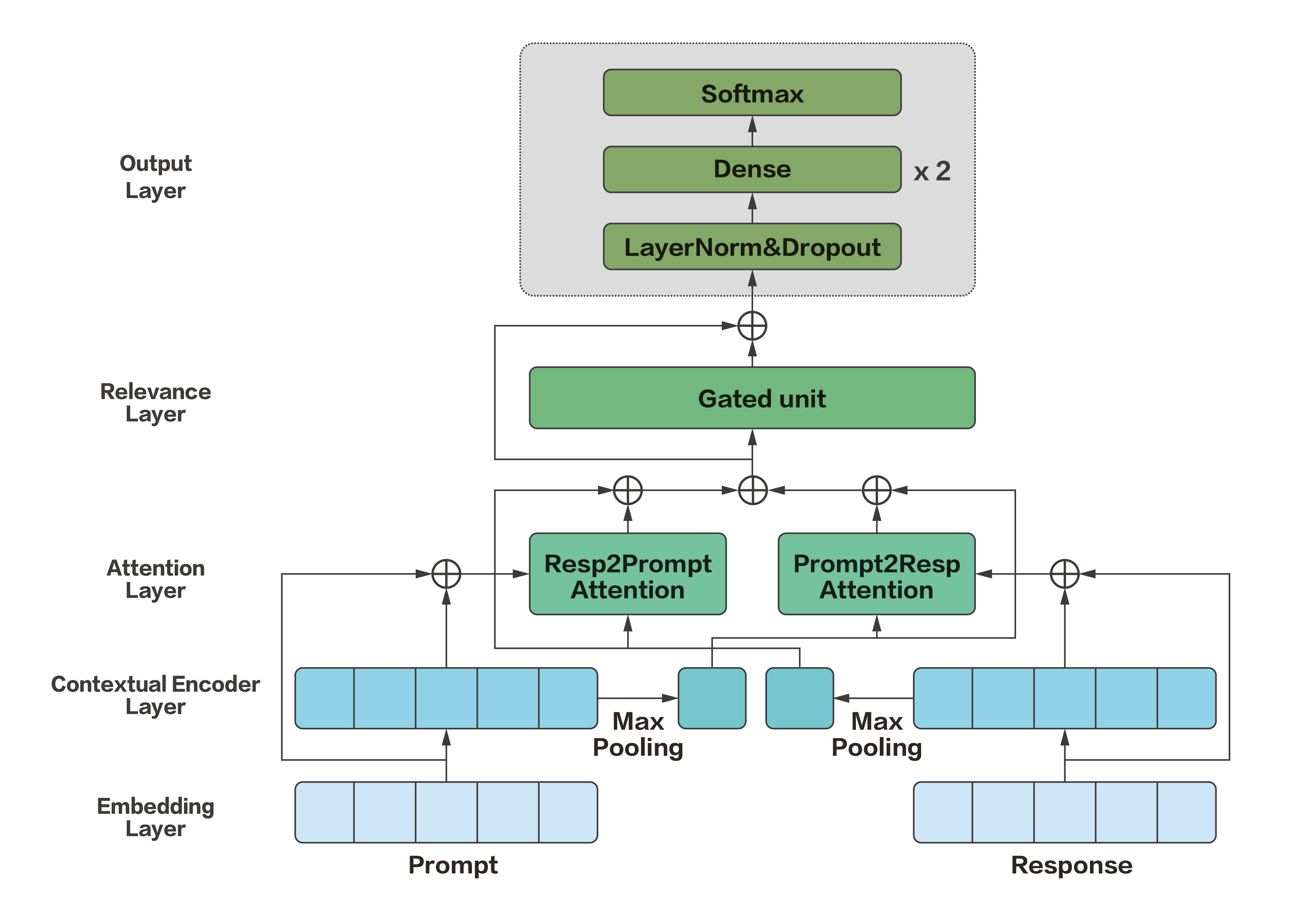}
\caption{An overview of GCBiA. Residual connections were widely used to connect each two-layer. The first two layers are applied to both prompt and response. Convolutions are used in contextual encoder layer and bi-attention mechanism is applied in attention layer. After calculating by the relevance layer with the gated unit, the relevance vector is then fed into the output layer which consists of the normalization layer, dropout, two fully connection layers and softmax.}
\label{fig:model}
\end{figure*}

\begin{itemize}
	\item \textbf{Word Embedding Layer} maps each word to a vector space using a pre-trained word embedding model.
    \item \textbf{Contextual Encoder Layer} utilizes contextual information from surrounding words to reinforce the embedding of the words. 
    These first two layers are applied to both prompts and responses.
    \item \textbf{Attention Layer} uses the attention mechanism in both directions, prompt-to-response and response-to-prompt, which provides complementary information to each other.
    \item \textbf{Relevance Layer} captures the important parts of the response regarding a prompt via the gated unit.
    \item \textbf{Output Layer} predicts whether the response is off-topic given the prompt.
\end{itemize}

In detail, each layer is illustrated as follows:
\begin{enumerate}
\item \textbf{Word Embedding Layer.} We first convert words to respective trainable word embeddings, initialized by pre-trained Glove~\citep{pennington2014glove}. 
The embeddings of prompts $W^P=\{w^P_t\}^n_{t=1}$ and responses $W^R=\{w^R_t\}^m_{t=1}$ are passed directly to the next contextual encoder layer. 

\item \textbf{Contextual Encoder Layer.} A stack of convolutional layers are employed to extract salient n-gram features from prompts and responses, aiming at creating an informative latent semantic representation of prompts and responses for the next layer. 
The $l$-th convolutional layer with one filter is represented as $c_i^l$ in Equation (\ref{eq:conv-layer1}), where $W\in\mathbb{R}^{k\times{d}}$, $b\in\mathbb{R}^d$.
We ensure that the output of each stack matches the input length by padding the input of each stack. 
The number of convolutional layers $l$ is 7, the kernel size $k$ is 7 and the number of filters in each convolutional layer is 128.
\begin{align}
\label{eq:conv-layer1}
c_i^l = f(W^l[c_{i-k/2}^{l-1},...,c_{i+k/2}^{l-1}] + b^l)
\end{align}
After the convolutional representation of prompts $U^P$ and responses $U^R$ in Equation (\ref{eq:conv-layer2s}-\ref{eq:conv-layer2e}) are obtained, a max pooling layer to extract the fixed-length vector is performed, seen in Equation (\ref{eq:conv-layer3s}-\ref{eq:conv-layer3e}). 
Max-pooling can keep the most salient n-gram features across the whole prompt/response.
\begin{align}
\label{eq:conv-layer2s}
U^P &= CONV(W^P) \\
\label{eq:conv-layer2e}
U^R &= CONV(W^R) \\
\label{eq:conv-layer3s}
v^P &= maxpooling(U^P) \\
\label{eq:conv-layer3e}
v^R &= maxpooling(U^R)
\end{align}
\item \textbf{Attention Layer.} In this layer, the attention mechanism is used in both directions, prompt-to-response and response-to-prompt, which provides complementary information to each other. 
However, unlike bi-attention applied to question answering and machine comprehension, including QANet~\citep{yu2018qanet}, BiDAF~\citep{seo2016bidirectional} and BiDAF++~\citep{choi2018quac}, we use max-pooling of CNN representation on prompt/response to summarize the prompt/response into a fixed-size vector.
\paragraph{Prompt-to-Response Attention.} Prompt-to-Response attention implicitly models which response words are more related to the whole prompt, which is crucial to assess the relevance of responses and prompts. 
Given max pooling vector $v^P$ of the prompt and CNN representation $U^R=\{u^R_t\}^m_{t=1}$ of the response, together with $W^P=\{w^P_t\}^n_{t=1}$ and $W^R=\{w^R_t\}^m_{t=1}$, Prompt-to-Response attention $c^R$ is then calculated in Equation (\ref{eq:attn-layer1s}-\ref{eq:attn-layer1e}), where the similarity function used is trilinear function~\citep{yu2018qanet} and residual connections are used.
\begin{gather}
\label{eq:attn-layer1s}
\tilde{u}^R_j = [u^R_j, w^R_j] \\
\tilde{v}^P =  [v^P, avgpooling(W^P)] \\
s_j = W[\tilde{u}^R_j, \tilde{v}^P, \tilde{u}^R_j \odot \tilde{v}^P] \\
\label{eq:attn-layer1m}
\alpha_i = \frac{exp(s_i)}{\sum_{j=1}^{m}exp(s_j)}  \\
\label{eq:attn-layer1e}
c^R = \sum^m_{i=1}{\alpha}_i\tilde{u}^R_i
\end{gather}
\paragraph{Response-to-Prompt Attention.} Similarly, Response-to-Prompt attention implicitly models which prompt words are more related to the whole response. 
The calculation of Response-to-Prompt attention, seen in Equation (\ref{eq:attn-layer2s}-\ref{eq:attn-layer2e}), is close to Prompt-to-Response attention.
\begin{gather}
\label{eq:attn-layer2s}
\tilde{u}^P_j = [u^P_j, w^P_j] \\
\tilde{v}^R =  [v^R, avgpooling(W^R)] \\
s_j = W[\tilde{u}^P_j, \tilde{v}^R, \tilde{u}^P_j \odot \tilde{v}^R] \\
\alpha_i = \frac{exp(s_i)}{\sum_{j=1}^{m}exp(s_j)}  \\
\label{eq:attn-layer2e}
c^P = \sum^n_{i=1}{\alpha}_i\tilde{u}^P_i
\end{gather}
\item \textbf{Relevance Layer.} To capture the important parts of responses and attend to the ones relevant to the prompts, we use one gated unit in this layer seen in Equation (\ref{eq:filter-layer1s}-\ref{eq:filter-layer1e}).
This gated unit focuses on the relation between the prompt and the response. Only relevant parts of each side can remain after the sigmoid operation.
The input of this layer is $({\tilde{c}}^R=[c^R,v^R], {\tilde{c}}^P=[c^P,v^P])$, which uses residual connections of the previous two layers.
\begin{gather}
\label{eq:filter-layer1s}
g = sigmoid(W_g[{\tilde{c}}^R, {\tilde{c}}^P]) \\
\label{eq:filter-layer1e}
[{\tilde{c}}^R, {\tilde{c}}^P]^* = g \odot [{\tilde{c}}^R, {\tilde{c}}^P]
\end{gather}
\item \textbf{Output Layer.} The fixed-length semantic matching vector produced by the previous layer and the previous second layer vector, are fed into the last output layer. 
It consists of one normalization layer, one dropout, two fully connected layers, and one softmax layer. 
The output distribution indicates the relevance of the prompt and the response.
We classify the output into two categories on-topic or off-topic through the threshold. 
Different threshold is chosen for the different prompt to make sure the on-topic recall of the prompt meets the lowest requirement, such as 0.999 for the online product system in our study.
\end{enumerate}

\section{Data}
\subsection{Dataset}
\label{sec:dataset}
Data from our IELTS speaking test mobile app\footnote{https://www.liulishuo.com/ielts.html} was used for training and testing in this paper. 
There are three parts in the IELTS\footnote{https://www.ielts.org/about-the-test/test-format} test:
Part1 focuses on general questions about test-takers and a range of familiar topics, such as home, family, work, studies, and interests.
In Part2, test-takers will be asked to talk about a particular topic. Discussion of more abstract ideas and issues about Part2 will occur in Part3.
Here is an example from our IELTS speaking test mobile app, seen in Table~\ref{tab:dataset_example}.
\begin{table}
\centering
\begin{tabular}{lp{6cm}}
\hline
\textbf{Part} & \textbf{Prompt}\\
\hline
Part1 & How long have you lived in your hometown? \\
Part2 & Describe something you bought according to an advertisement you saw. what it was where you saw or heard about it what it was about. \\
Part3 & Do you trust advertisements? \\
\hline
\end{tabular}
\caption{An example from our IELTS speaking test mobile app.}\label{tab:dataset_example}
\end{table}

All responses from test-takers were generated from our automatic speech recognition (ASR) system, which will be briefly introduced in Section~\ref{sec:asr}. 
Responses for a target prompt collected in our paid service were used as its on-topic training examples, and responses from the other prompts were used as the off-topic training examples for the target prompt.
It is a reasonable setup because most of the responses in our paid service are on-topic (we labeled about 5K responses collected under our paid service and found only 1.3\% of them are off-topic) and a certain level of “noise” in the training is acceptable.
The test data was produced in the same way as the training data except that human validation was further introduced to ensure its validity.
To ensure the authenticity of our train and test data further, we filter short responses for each part. 
The length of words from each response in Part1, Part2, and Part3 should be over 15, 50, and 15, respectively. 

Table~\ref{tab:dataset} shows the details of our train and test datasets:
1.12M responses from 1356 prompts are used to train our model.
The average number of responses to each prompt is 822.
The number of on-topic and off-topic responses are 564.3K and 551.3K in training data. 
We divide the test data into two parts: seen benchmark and unseen benchmark.
Prompts of the seen benchmark can appear in train data, while prompts of unseen benchmark cannot. 
The seen benchmark consists of 33.6K responses from 156 prompts, including 17.7K on-topic responses and 15.9K off-topic responses, and the average number of responses of each prompt is 216.
In the unseen benchmark, there are 10.1K responses from 50 prompts, including 5.0K on-topic responses and 5.1K off-topic responses, and the average number of responses of each prompt is 202.
\begin{table*}[h!]
\centering
\begin{tabular}{ll|ccccc}
\hline
Data & &\#Prompt &\#Resp. &\#Resp./Prompt &On-topic &Off-topic \\
\hline
Train & &1356 &1,12M &822 &564.3K &551.3K \\
\multirow{2}{*}{Test}&Seen &156 &33.6K &216 &17.7K &15.9K \\
&Unseen &50 &10.1K &202 &5.0K &5.1K \\
\hline
\end{tabular}
\caption{The train and test datasets for off-topic detection task}
\label{tab:dataset}
\end{table*}

\subsection{ASR System}
\label{sec:asr}
A hybrid deep neural network DNN-HMM system is used for ASR.
The acoustic model contains 17 sub-sampled time-delay neural network layers with low-rank matrix factorization (TDNNF)~\citep{povey2018semi}, and is trained on over 8000 hours of speech, using the lattice-free MMI~\citep{povey2016purely} recipe in Kaldi\footnote{http://kaldi-asr.org} toolkit. 
A tri-gram LM with Kneser-Ney smoothing is trained using the SRILM\footnote{http://www.speech.sri.com/projects/srilm/} toolkit and applied at first pass decoding to generate word lattices.
An RNN-LM~\citep{mikolov2010recurrent} is applied to re-scoring the lattices to achieve the final recognition results. 
The ASR system achieves a word error rate of around 13\% on our 50 hours ASR test set.

\subsection{Metric}
We use two assessment metrics in this paper:
Average Off-topic Recall (AOR) and Prompt Ratio over Recall0.3 (PRR3).
AOR denotes the average number of off-topic responses recall of all prompts (156 prompts on the seen benchmark and 50 prompts on the unseen benchmark).
PRR3 denotes the ratio of prompts whose off-topic recall is over 0.3.

Here is a case of AOR and PRR3 on seen benchmark: three prompts have 102, 102, and 102 off-topic responses, respectively.
Suppose that we have recalled 100, 90 and 30 off-topic responses for the three prompts, off-topic recall of each prompt is 100/102=98.0\%, 90/102=88.2\%, and 30/102=29.4\%. 
In this case AOR=(100/102 + 90/102 + 30/102)/3=71.9\%, and PRR3=2/3=66.7\%.
To ensure that the off-topic detection is applicable in real scenes, high on-topic recall (0.999 in this paper) is required. 
We give restriction that the on-topic recall on each prompt should be over 0.999 when calculating AOR and PRR3.

\subsection{Training settings}
The model is implemented by Keras\footnote{https://keras.io/}. 
We use pre-trained Glove as word embedding, the dimension of which is 300.
The train and dev batch size are 1024 and 512. 
The kernel size, filter number, and block size of CNN are 7, 128, and 7 by tuning on the dev set.
The fix-length of prompts and responses are 40 and 280 according to the length distribution of prompts and responses in the training data.
Nadam~\citep{dozat2016incorporating} is used as our optimizer with a learning rate of 0.002.
The loss function is binary cross-entropy.
The epoch size is 20, and we apply early-stop when dev loss has not been improving for three epochs.
Our GCBiA model and inference code is released here.\footnote{https://github.com/zhayefei/off-topic-GCBiA}

\section{Experiments}
\subsection{Results}
\label{sec:result}
\begin{table*}[h!]
\centering
\begin{tabular}{l|l||cc|cc}
\hline
\multirow{2}*{Systems} &\multirow{2}*{Model} &\multicolumn{2}{c}{Seen} &\multicolumn{2}{c}{Unseen} \\
\cline{3-6}
&&PRR3 &AOR &PRR3 &AOR \\
\hline
\citealp{malinin2017attention} &Att-RNN &84.6 &72.2 &32.0 &21.0 \\
\hline
Our baseline model &G-Att-RNN &87.8 &76.8 &54.0 &38.1 \\
\hline
\multirow{4}{*}{This work} & ~~+ Bi-Attention &90.4  &78.3 &56.0 &39.7 \\
 		  				   & ~~~~ + RNN$\to$CNN	&89.7  &76.6 &66.0 	&43.7 \\
 		  				   & ~~~~~~ + $maxpooling$ &92.3  &79.1 &68.0  &42.2 \\
 		 				   & ~~~~~~~~ + Res-conn in gated unit (GCBiA) &\textbf{93.6} &\textbf{79.2} &\textbf{68.0} &\textbf{45.0}  \\
\hline
\end{tabular}
\caption{The comparison of different models based on over 0.999 on-topic recall on seen and unseen benchmarks. AOR means Average Off-topic Recall (\%) and PRR3 means Prompt Ratio over off-topic Recall 0.3 (\%).}
\label{tab:results}
\end{table*}
We carried out experiments on both seen benchmark and unseen benchmark mentioned in Section~\ref{sec:dataset}.
As is shown in Table~\ref{tab:results}, Att-RNN is our baseline model. 
To make the evaluation more convincing, we built a stronger baseline model G-Att-RNN based on Att-RNN by adding residual connections with each layer. 
Additionally, we add a gated unit as the relevance layer for our baseline model G-Att-RNN. 
Compared with Att-RNN, our baseline model G-Att-RNN achieved significant improvements on both seen (by +3.2 PRR3 points and +4.6 AOR points) and unseen benchmark (by +22.0 PRR3 points and +17.1 AOR). 

From Table~\ref{tab:results}, comparing with Att-RNN baseline, we can see that our approach GCBiA can achieve impressive improvements by +36.0 PRR3 points and +24.0 AOR points on the unseen benchmark, as well as +9.0 PRR3 points and +7.0 AOR points on the seen benchmark. 
Meanwhile, our approach significantly outperforms G-Att-RNN by +14.0 PRR3 points and + 6.9 AOR points on the unseen benchmark, as well as +5.8 PRR3 points and +2.4 AOR points on the seen benchmark.

\subsection{Ablation Studies}
\label{sec:ablation}
As gated unit and residual connections have been proved useful in Section~\ref{sec:result}, we conducted ablation analysis on seen and unseen benchmarks, seen in Table~\ref{tab:results}, to further study how other components contribute to the performance based on G-Att-RNN.

Because topic words of the prompt were focused on, the bi-attention mechanism is beneficial to replace the uni-attention by adding response-to-prompt attention, with +2.0 PRR3 points and +1.6 AOR points improvements on the unseen benchmark, as well as +2.6 PRR3 points and +1.5 AOR points on the seen benchmark. Besides, CNN with average-pooling applied to substitute RNN is also useful on the unseen benchmark by +10.0 PRR3 and +4.0 AOR points improvement. 
Though a little drop (-1.7\% on seen AOR) in performance was caused by CNN with average-pooling, CNN with max-pooling can achieve improvements on the seen benchmark by +2.6 PRR3 and + 2.5 AOR in return. 
In general, CNN is more suitable than RNN for the contextual encoder layer in our model framework, for seen and unseen prompts.
Finally, we also benefit from the residual connections for the gated unit with +2.8 AOR points improvement on the unseen benchmark.

\subsection{Analysis}
In this section, we analyzed the essence of our model from two perspectives. One is the bi-attention mechanism visualization and the other is the dimension reduction analysis of the semantic matching representation. More details are illustrated as follows: 

\begin{figure}[h!]
  \centering
  \begin{subfigure}[b]{1\linewidth}
    \includegraphics[width=\linewidth]{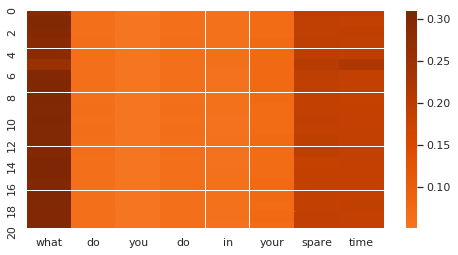}
     \caption{Attention on the prompt.}
    \label{fig:attn_heatmap:a}
  \end{subfigure}
  \begin{subfigure}[b]{1\linewidth}
    \includegraphics[width=\linewidth]{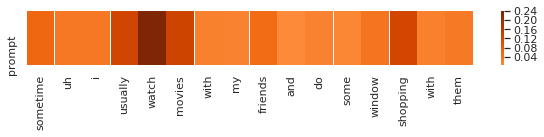}
    \caption{Attention on the on-topic response.}
    \label{fig:attn_heatmap:b}
  \end{subfigure}
   \begin{subfigure}[b]{1\linewidth}
    \includegraphics[width=\linewidth]{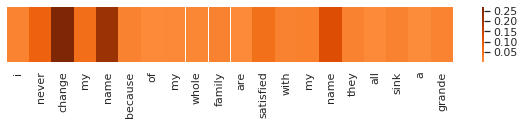}
    \caption{Attention on the off-topic response.}
    \label{fig:attn_heatmap:c}
  \end{subfigure}
  \caption{The heatmap of attention on the prompt and response.}
  \label{fig:attn_heatmap}
\end{figure}
\paragraph{Bi-Attention Visualization.} Figure~\ref{fig:attn_heatmap} gives the visualization of the bi-attention mechanism. 
Bi-attention mechanism can capture the interrogative ``what'' and topic words ``spare time'' of prompt ``what do you do in your spare time'' seen in subfigure~\ref{fig:attn_heatmap:a} , capture the key-phrases ``usually watch movies'' and ``shopping'' of the response seen in subfigure~\ref{fig:attn_heatmap:b}, and capture the key-phrases ``change name'' and ``name'' seen in subfigure~\ref{fig:attn_heatmap:c}.
Due to the increased focus on the prompt, bi-attention is more beneficial for assessing the relevance of responses and prompts by matching the key phrases or words between them.
The response in subfigure~\ref{fig:attn_heatmap:b} is classified as on-topic, while the response in subfigure~\ref{fig:attn_heatmap:c} is classified as off-topic.
 
\begin{figure}[h!]
  \centering
  \begin{subfigure}[b]{0.45\linewidth}
    \includegraphics[width=\linewidth]{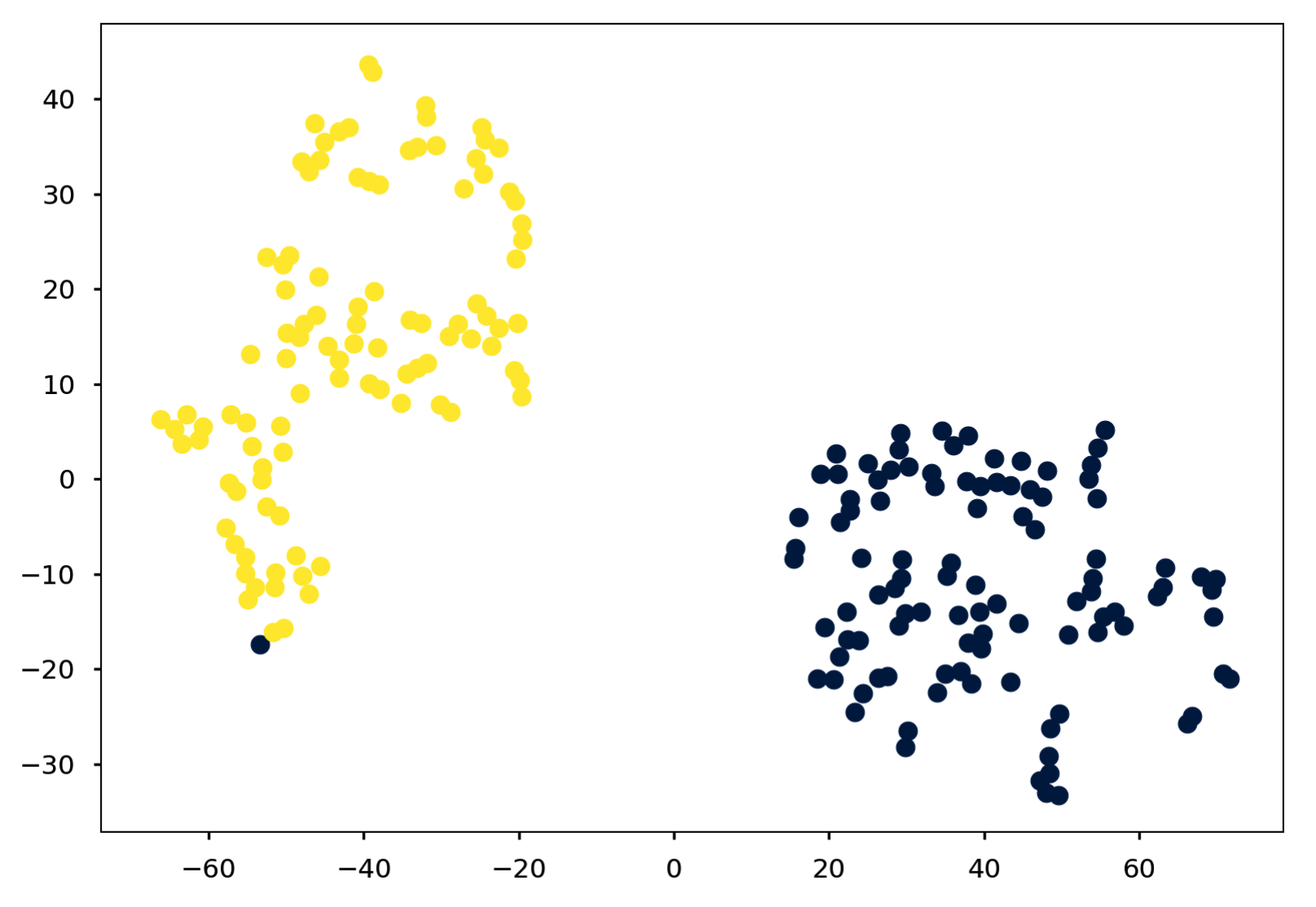}
     \caption{True resp distribution on clear-semantic topic prompt.}
     \label{fig:classify_prompts:a}
  \end{subfigure}
  \begin{subfigure}[b]{0.45\linewidth}
    \includegraphics[width=\linewidth]{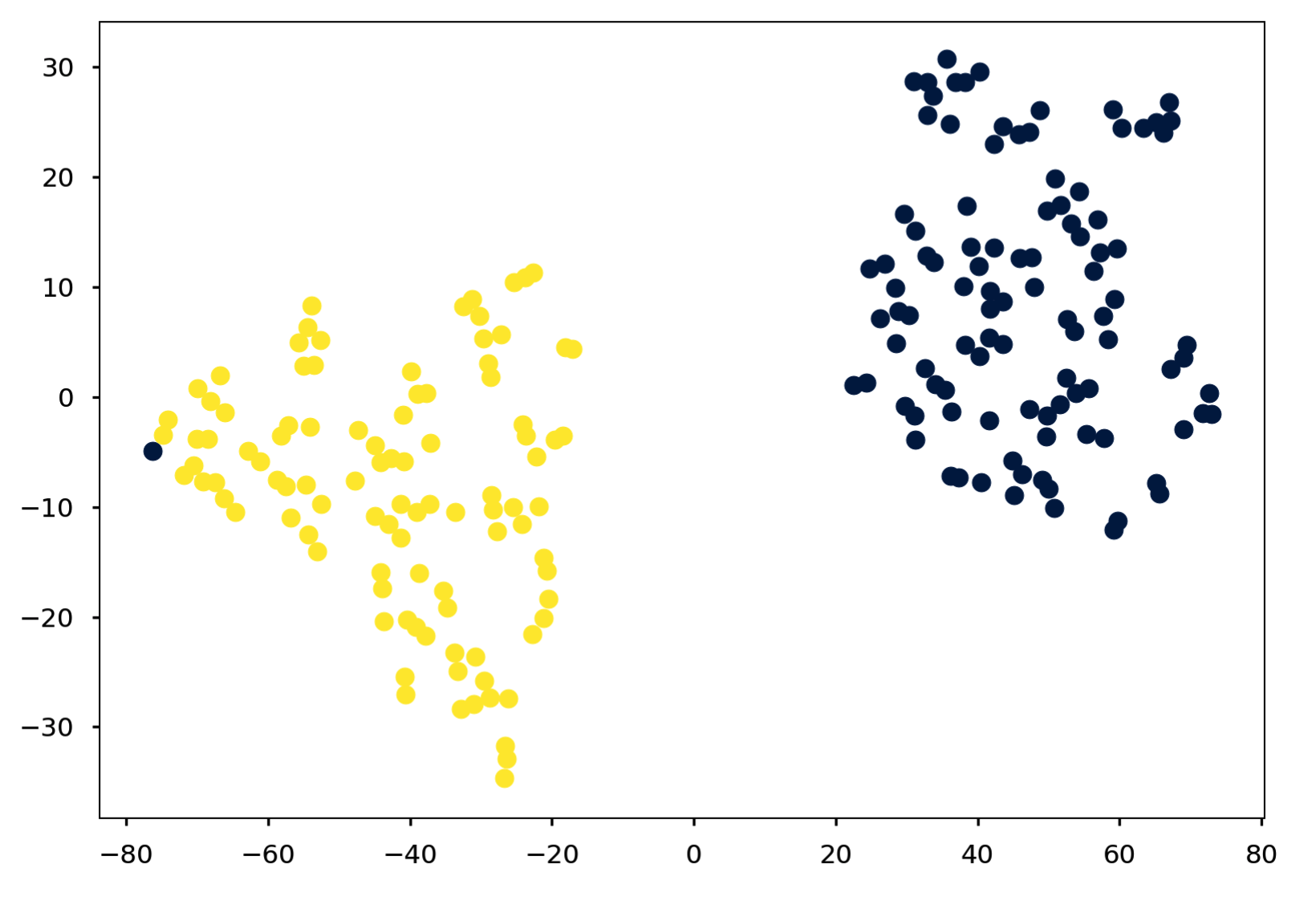}
    \caption{Model's resp distribution on clear-semantic topic prompt.}
    \label{fig:classify_prompts:b}
  \end{subfigure}
    \begin{subfigure}[b]{0.45\linewidth}
    \includegraphics[width=\linewidth]{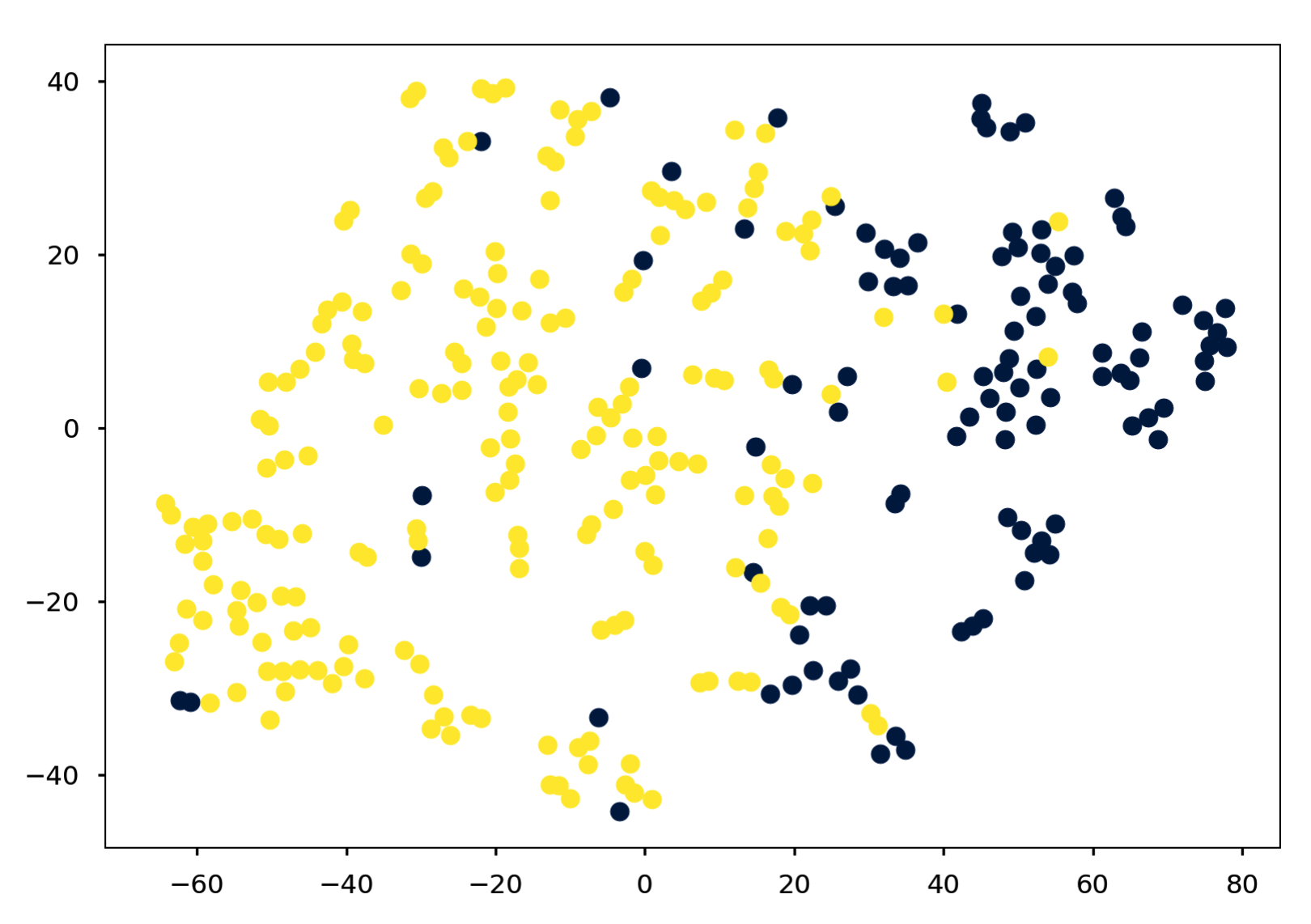}
     \caption{True response distribution on divergent prompt.}
     \label{fig:classify_prompts:c}
  \end{subfigure}
  \begin{subfigure}[b]{0.45\linewidth}
    \includegraphics[width=\linewidth]{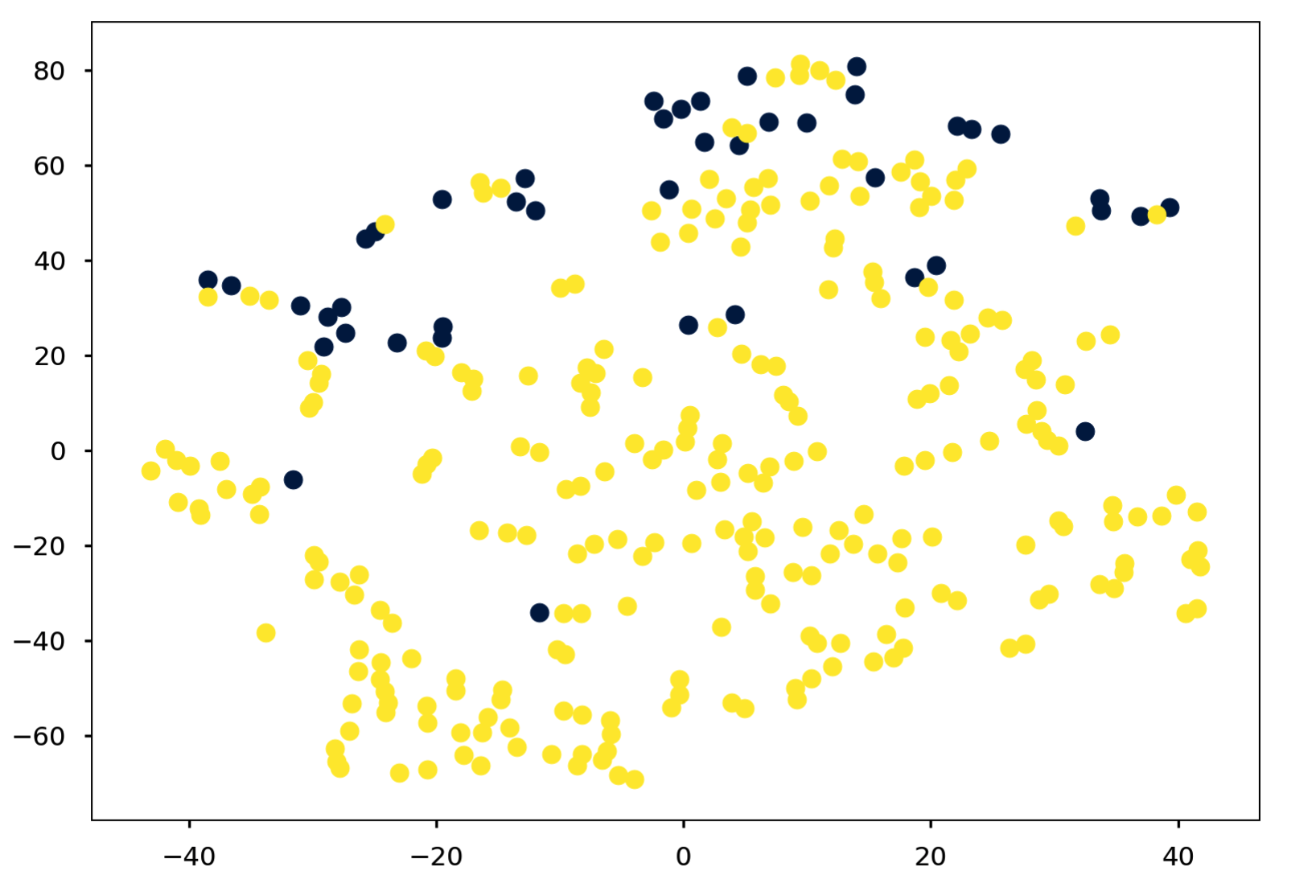}
    \caption{Model's resp distribution on divergent prompt.}
    \label{fig:classify_prompts:d}
  \end{subfigure}
  \caption{The analysis of response distribution on different types of prompts. The yellow and black colours represent the on-topic and off-topic response results respectively.}
  \label{fig:classify_prompts}
\end{figure}

\paragraph{Semantic Matching Representation Visualization.} As the output vector of the relevance layer using the gated unit can better represent the relevance of prompts and responses, the semantic matching representation was obtained from the relevance layer. 
With the help of t-SNE~\citep{maaten2008visualizing}, the visualization result was shown in Figure~\ref{fig:classify_prompts}. 
Subfigure~\ref{fig:classify_prompts:a} tells the true response distribution of one prompt, ``describe a special meal that you have had, what the meal was, who you had this meal with and explain why this meal was special'', which has a clear-semantic topic ``meal''.
Meanwhile, subfigure~\ref{fig:classify_prompts:b} reveals the response distribution using our semantic matching representation on the same prompt as subfigure~\ref{fig:classify_prompts:a} .

We can see that semantic matching representation of our model maintains good performance on this kind of prompt, which has one clear-semantic topic to limit the discussion in one scope.
Additionally, some prompts are open to discuss, which are divergent. 
Given a case of the prompt ``what do you do in your spare time'', and we can observe its true response distribution in subfigure~\ref{fig:classify_prompts:c} . 
Compared with it in subfigure~\ref{fig:classify_prompts:c} , our model tends to predict responses on-topic, seen in subfigure~\ref{fig:classify_prompts:d} , because high on-topic recall (0.999) is limited.

\subsection{Negative Sampling Augmentation Method}

\begin{figure}[h!]
  \centering
 	\includegraphics[width=8cm,height=6cm]{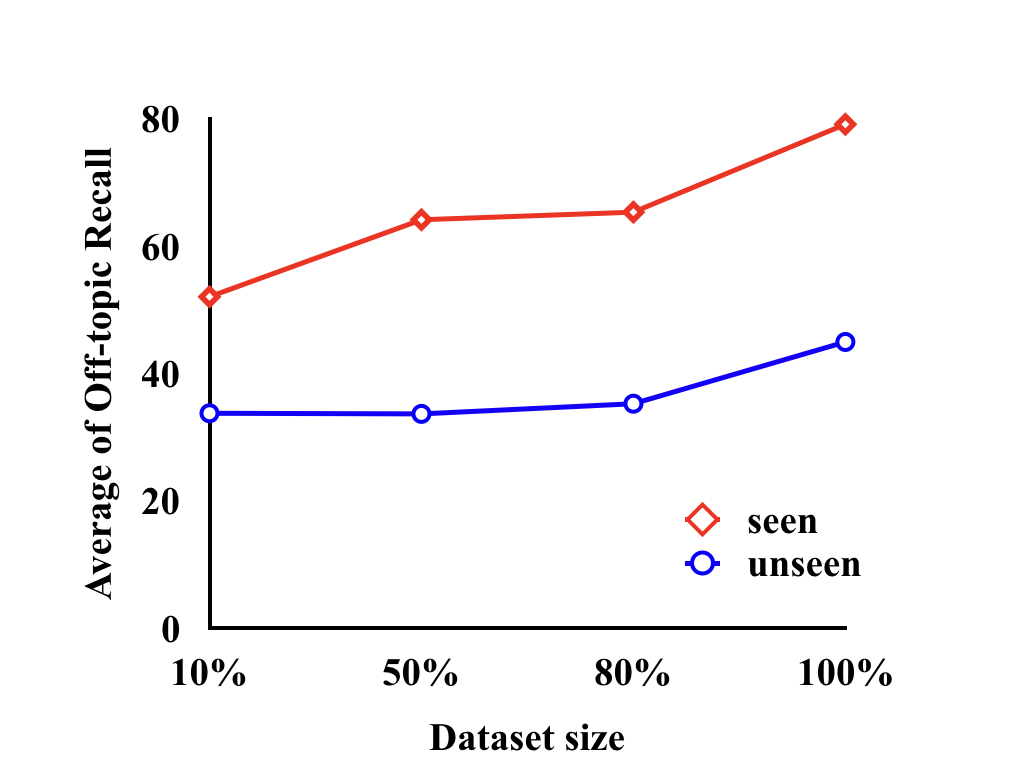}
    \caption{Trends of AOR (Average Off-topic Recall) on seen and unseen prompts with datasize variation.}
  \label{fig:datasize_variation}
\end{figure}
To investigate the impact of training data size, we conduct some experiments with varying sizes of training data. 
In figure~\ref{fig:datasize_variation}, we find that the larger the training data size, the better the performance. 

\begin{table}[h!]
\centering
\begin{tabular}{p{2.3cm}|cc|cc}
\hline
\multirow{2}*{Model} &\multicolumn{2}{c}{Seen} &\multicolumn{2}{c}{Unseen} \\
\cline{2-5}
&PRR3 &AOR &PRR3 &AOR \\
\hline
GCBiA &93.6 &79.2 &68.0 &45.0 \\
\hline
+ neg sampling&\textbf{94.2} &\textbf{88.2} &\textbf{79.4} &\textbf{69.1} \\
\hline
\end{tabular}
\caption{The performance of GCBiA with negative sampling augmentation method conditioned on over 0.999 on-topic recall.}
\label{tab:neg_samples}
\end{table}

To augment training data and strengthen the generalization of the off-topic response detection model for unseen prompts, we proposed a new and effective negative sampling method for off-topic response detection task.
Comparing with the previous method of generating only one negative sample for each positive one, we generated two. 
The first one is chosen randomly as before, and the second one consists of words shuffled from the first one. 
This method contributes to the diversity of negative samples of training data.
The size of our training data reaches 1.67M, compared with 1.12M in the previous negative sampling method. 
To make training data balanced, we gave the weight of positive and negative samples: 1 and 0.5, respectively.
As is shown in Table~\ref{tab:neg_samples}, a significant performance improvement (+9.0 seen AOR and +24.1 unseen AOR) is achieved by this negative sampling method.
Our model GCBiA equipped with negative sampling augmentation can achieve 88.2\% and 69.1\% average off-topic response recall on seen and unseen prompts, conditioned on 0.999 on-topic recall.

\section{Conclusion}
In this paper, we conducted a series of work around the task of off-topic response detection.
First of all, a model framework of five major layers was proposed, within which bi-attention mechanism and convolutions were used to well capture the topic words of prompts and key-phrase of responses, and gated unit as relevance layer was applied to better obtaining semantic matching representation, as well as residual connections with each major layer.
Moreover, the visualization analysis of the off-topic model was given to study the essence of the model.
Finally, a novel negative sampling augmentation method was introduced to augment off-topic training data.
We verified the effectiveness of our approach and achieved significant improvements on both seen and unseen test data.

\section*{Acknowledgments}
We are grateful to our colleague Bin Wang for helping with the ASR system. We thank our colleague Puyu Chen for proofreading. Last but not least, we thank the anonymous reviewers for their invaluable comments.

\bibliography{anthology,acl2020}

\begin{thebibliography}{22}
\expandafter\ifx\csname natexlab\endcsname\relax\def\natexlab#1{#1}\fi

\bibitem[{Choi et~al.(2018)Choi, He, Iyyer, Yatskar, Yih, Choi, Liang, and
  Zettlemoyer}]{choi2018quac}
Eunsol Choi, He~He, Mohit Iyyer, Mark Yatskar, Wen-tau Yih, Yejin Choi, Percy
  Liang, and Luke Zettlemoyer. 2018.
\newblock Quac: Question answering in context.
\newblock \emph{arXiv preprint arXiv:1808.07036}.

\bibitem[{Dozat(2016)}]{dozat2016incorporating}
Timothy Dozat. 2016.
\newblock Incorporating nesterov momentum into adam.

\bibitem[{Evanini and Wang(2014)}]{evanini2014automatic}
Keelan Evanini and Xinhao Wang. 2014.
\newblock Automatic detection of plagiarized spoken responses.
\newblock In \emph{Proceedings of the Ninth Workshop on Innovative Use of NLP
  for Building Educational Applications}, pages 22--27.

\bibitem[{He et~al.(2016)He, Zhang, Ren, and Sun}]{he2016deep}
Kaiming He, Xiangyu Zhang, Shaoqing Ren, and Jian Sun. 2016.
\newblock Deep residual learning for image recognition.
\newblock In \emph{Proceedings of the IEEE conference on computer vision and
  pattern recognition}, pages 770--778.

\bibitem[{Higgins and Heilman(2014)}]{higgins2014managing}
Derrick Higgins and Michael Heilman. 2014.
\newblock Managing what we can measure: Quantifying the susceptibility of
  automated scoring systems to gaming behavior.
\newblock \emph{Educational Measurement: Issues and Practice}, 33(3):36--46.

\bibitem[{Lee et~al.(2017)Lee, Yoon, Wang, Mulholland, Choi, and
  Evanini}]{lee2017off}
Chong~Min Lee, Su-Youn Yoon, Xihao Wang, Matthew Mulholland, Ikkyu Choi, and
  Keelan Evanini. 2017.
\newblock Off-topic spoken response detection using siamese convolutional
  neural networks.
\newblock In \emph{INTERSPEECH}, pages 1427--1431.

\bibitem[{Lochbaum et~al.(2013)Lochbaum, Rosenstein, Foltz, Derr
  et~al.}]{lochbaum2013detection}
Karen~E Lochbaum, Mark Rosenstein, PW~Foltz, Marcia~A Derr, et~al. 2013.
\newblock Detection of gaming in automated scoring of essays with the iea.
\newblock In \emph{National Council on Measurement in Education Conference
  (NCME), San Francisco, CA}.

\bibitem[{Louis and Higgins(2010)}]{louis2010off}
Annie Louis and Derrick Higgins. 2010.
\newblock Off-topic essay detection using short prompt texts.
\newblock In \emph{proceedings of the NAACL HLT 2010 fifth workshop on
  innovative use of NLP for building educational applications}, pages 92--95.
  Association for Computational Linguistics.

\bibitem[{Maaten and Hinton(2008)}]{maaten2008visualizing}
Laurens van~der Maaten and Geoffrey Hinton. 2008.
\newblock Visualizing data using t-sne.
\newblock \emph{Journal of machine learning research}, 9(Nov):2579--2605.

\bibitem[{Malinin et~al.(2017)Malinin, Knill, Ragni, Wang, and
  Gales}]{malinin2017attention}
Andrey Malinin, Kate Knill, Anton Ragni, Yu~Wang, and Mark~JF Gales. 2017.
\newblock An attention based model for off-topic spontaneous spoken response
  detection: An initial study.
\newblock In \emph{SLaTE}, pages 144--149.

\bibitem[{Malinin et~al.(2016)Malinin, Van~Dalen, Wang, Knill, and
  Gales}]{malinin2016off}
Andrey Malinin, Rogier~C Van~Dalen, Yu~Wang, Katherine~Mary Knill, and
  Mark~John Gales. 2016.
\newblock Off-topic response detection for spontaneous spoken english
  assessment.

\bibitem[{Mikolov et~al.(2010)Mikolov, Karafi{\'a}t, Burget,
  {\v{C}}ernock{\`y}, and Khudanpur}]{mikolov2010recurrent}
Tom{\'a}{\v{s}} Mikolov, Martin Karafi{\'a}t, Luk{\'a}{\v{s}} Burget, Jan
  {\v{C}}ernock{\`y}, and Sanjeev Khudanpur. 2010.
\newblock Recurrent neural network based language model.
\newblock In \emph{Eleventh annual conference of the international speech
  communication association}.

\bibitem[{Pennington et~al.(2014)Pennington, Socher, and
  Manning}]{pennington2014glove}
Jeffrey Pennington, Richard Socher, and Christopher Manning. 2014.
\newblock Glove: Global vectors for word representation.
\newblock In \emph{Proceedings of the 2014 conference on empirical methods in
  natural language processing (EMNLP)}, pages 1532--1543.

\bibitem[{Povey et~al.(2018)Povey, Cheng, Wang, Li, Xu, Yarmohammadi, and
  Khudanpur}]{povey2018semi}
Daniel Povey, Gaofeng Cheng, Yiming Wang, Ke~Li, Hainan Xu, Mahsa Yarmohammadi,
  and Sanjeev Khudanpur. 2018.
\newblock Semi-orthogonal low-rank matrix factorization for deep neural
  networks.
\newblock In \emph{Interspeech}, pages 3743--3747.

\bibitem[{Povey et~al.(2016)Povey, Peddinti, Galvez, Ghahremani, Manohar, Na,
  Wang, and Khudanpur}]{povey2016purely}
Daniel Povey, Vijayaditya Peddinti, Daniel Galvez, Pegah Ghahremani, Vimal
  Manohar, Xingyu Na, Yiming Wang, and Sanjeev Khudanpur. 2016.
\newblock Purely sequence-trained neural networks for asr based on lattice-free
  mmi.
\newblock In \emph{Interspeech}, pages 2751--2755.

\bibitem[{Seo et~al.(2016)Seo, Kembhavi, Farhadi, and
  Hajishirzi}]{seo2016bidirectional}
Minjoon Seo, Aniruddha Kembhavi, Ali Farhadi, and Hannaneh Hajishirzi. 2016.
\newblock Bidirectional attention flow for machine comprehension.
\newblock \emph{arXiv preprint arXiv:1611.01603}.

\bibitem[{Wang et~al.(2017)Wang, Yang, Wei, Chang, and Zhou}]{wang2017gated}
Wenhui Wang, Nan Yang, Furu Wei, Baobao Chang, and Ming Zhou. 2017.
\newblock Gated self-matching networks for reading comprehension and question
  answering.
\newblock In \emph{Proceedings of the 55th Annual Meeting of the Association
  for Computational Linguistics (Volume 1: Long Papers)}, pages 189--198.

\bibitem[{Wang et~al.(2019)Wang, Yoon, Evanini, Zechner, and
  Qian}]{wang2019automatic}
Xinhao Wang, Su-Youn Yoon, Keelan Evanini, Klaus Zechner, and Yao Qian. 2019.
\newblock Automatic detection of off-topic spoken responses using very deep
  convolutional neural networks.
\newblock \emph{Proc. Interspeech 2019}, pages 4200--4204.

\bibitem[{Yin et~al.(2017)Yin, Kann, Yu, and Sch{\"u}tze}]{yin2017comparative}
Wenpeng Yin, Katharina Kann, Mo~Yu, and Hinrich Sch{\"u}tze. 2017.
\newblock Comparative study of cnn and rnn for natural language processing.
\newblock \emph{arXiv preprint arXiv:1702.01923}.

\bibitem[{Yoon and Xie(2014)}]{yoon2014similarity}
Su-Youn Yoon and Shasha Xie. 2014.
\newblock Similarity-based non-scorable response detection for automated speech
  scoring.
\newblock In \emph{Proceedings of the Ninth Workshop on Innovative Use of NLP
  for Building Educational Applications}, pages 116--123.

\bibitem[{Young et~al.(2018)Young, Hazarika, Poria, and
  Cambria}]{young2018recent}
Tom Young, Devamanyu Hazarika, Soujanya Poria, and Erik Cambria. 2018.
\newblock Recent trends in deep learning based natural language processing.
\newblock \emph{ieee Computational intelligenCe magazine}, 13(3):55--75.

\bibitem[{Yu et~al.(2018)Yu, Dohan, Luong, Zhao, Chen, Norouzi, and
  Le}]{yu2018qanet}
Adams~Wei Yu, David Dohan, Minh-Thang Luong, Rui Zhao, Kai Chen, Mohammad
  Norouzi, and Quoc~V Le. 2018.
\newblock Qanet: Combining local convolution with global self-attention for
  reading comprehension.
\newblock \emph{arXiv preprint arXiv:1804.09541}.

\end{thebibliography}
\bibliographystyle{acl_natbib}

\end{document}